\documentclass[11pt,a4paper]{article}
\usepackage{natbib}
\usepackage{graphicx}
\usepackage{wrapfig}
\usepackage[labelfont=bf]{caption}
\usepackage{amsmath}
\usepackage{amsfonts}
\usepackage{amssymb}
\usepackage{makeidx}
\usepackage{setspace}
\usepackage{geometry}
\usepackage{color}
\usepackage{hyperref}
\usepackage{xcolor}
\usepackage{epigraph}
\hypersetup{colorlinks, linkcolor={red!50!black}, citecolor={blue!50!black}, urlcolor={blue!80!black}}
\title{HVM-1: Large-scale video models pretrained with nearly 5000 hours of human-like video data}
\author{
  Emin Orhan \\
  \texttt{aeminorhan@gmail.com}
  }
\date{}

\begin{document}

\maketitle

\begin{abstract}
We introduce Human-like Video Models (HVM-1), large-scale video models pretrained with nearly 5000 hours of curated human-like video data (mostly egocentric, temporally extended, continuous video recordings), using the spatiotemporal masked autoencoder (ST-MAE) algorithm. We release two 633M parameter models trained at spatial resolutions of 224$\times$224 and 448$\times$448 pixels. We evaluate the performance of these models in downstream few-shot video and image recognition tasks and compare them against a model pretrained with 1330 hours of short action-oriented video clips from YouTube (Kinetics-700). HVM-1 models perform competitively against the Kinetics-700 pretrained model in downstream evaluations despite substantial qualitative differences between the spatiotemporal characteristics of the corresponding pretraining datasets. HVM-1 models also learn more accurate and more robust object representations compared to models pretrained with the image-based MAE algorithm on the same data, demonstrating the potential benefits of learning to predict temporal regularities in natural videos for learning better object representations.
\end{abstract}

\section{Introduction}
\epigraph{... but \textit{time and chance} happeneth to them all.}{---Ecclesiastes 9:11}
\vspace{-1.4em}

Modern machine learning models are often trained with large amounts of data that is both qualitatively and quantitatively very different from the kind of data humans learn from during their development. For example, it is not uncommon for language models to be trained on trillions of tokens of text or for vision models to be trained on millions of static images scraped from the internet and from other sources. Humans, on the other hand, acquire their linguistic capabilities from about 100 million tokens at most, and their visual capabilities from a few years of continuous, egocentric video stream. This prominent ``data gap'' \citep{frank2023} has led to a growing interest in understanding what exactly modern machine learning algorithms can learn from both qualitatively and quantitatively more human-like data \citep{bambach2018,zhuang2021,zaadnoordijk2022,warstadt2023,orhan2024,vong2024,sheybani2024}. Understanding this would have two main benefits. First, it would reveal to what extent current machine learning algorithms \textit{require} the type of unhuman-like data they are typically trained with today in order to achieve human-level, or beyond human-level, capabilities (in other words, to what extent their success depends on the type of data they are typically trained with today). Secondly, it would also shed light on the long-standing nature \textit{vs.}~nurture question in developmental psychology \citep{wood2024}. Since modern machine learning algorithms are, by and large, extremely generic algorithms, understanding what they can learn from human-like data would inform us about the capabilities that can be learned from such data without strong innate constraints or inductive biases.

To contribute to this recent effort to understand what modern machine learning algorithms can learn from more human-like data, here we introduce \textbf{H}uman-like \textbf{V}ideo \textbf{M}odels (HVM-1), large-scale video models pretrained with nearly 5000 hours of curated human-like video data and evaluate the capabilities of the models in downstream tasks. We believe HVM-1 models are the largest and the most capable video models to date trained with human-like video data and we hope that they will be a useful resource particularly for researchers working at the intersection of machine learning and cognitive science. Code, models, and various tools to use and analyze the models are available from the following public repository: \href{https://github.com/eminorhan/hvm-1}{https://github.com/eminorhan/hvm-1}.

\section{Methods}
\subsection{Pretraining data}
\begin{figure}
  \centering
	\includegraphics[width=1\textwidth, trim=2mm 2mm 2mm 2mm, clip]{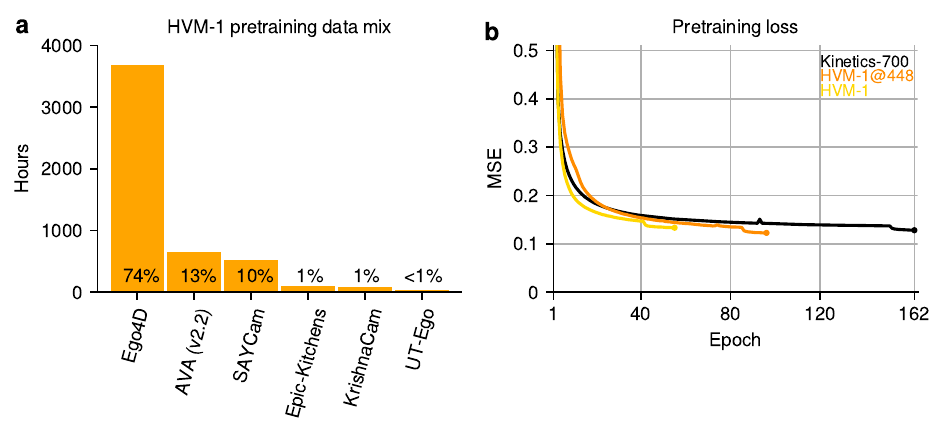} 
	\caption{\textbf{(a)} Datasets used for pretraining HVM-1 models. \textbf{(b)} Evolution of the training loss for the three pretrained models over the course of pretraining.}
  \label{pretraining_fig}
\end{figure}

\textbf{HVM-1 pretraining data.} We use a combination of six human-like video datasets previously described in \cite{orhan2023}. Briefly, the combined dataset consists of the following individual components (Figure~\ref{pretraining_fig}a): Ego4D \citep{grauman2022}, AVA v2.2 \citep{gu2018}, SAYCam \citep{sullivan2021}, Epic-Kitchens \citep{damen2018}, KrishnaCam \citep{singh2016}, and UT Ego \citep{lee2012}. We refer the reader to \cite{orhan2023} for a more detailed description of these datasets. The videos in these datasets are \textit{human-like} in two important respects: (i) they consist of mostly egocentric, natural videos recorded from the perspective of people while performing daily activities and (ii) they are continuous, temporally extended videos lasting on the order of tens of minutes to hours in duration. The combined dataset contains \colorbox{lime}{4971 hours} of such human-like video in total and the Ego4D dataset makes up the bulk of the data (74\% of the total data; see Figure~\ref{pretraining_fig}a). To make data loading efficient, we divide the entire dataset into approximately 104K individual segments (video files) with an average duration of 2.9 minutes per segment. We then remove any blank or corrupted segments that may have occurred due to camera failures in the original recordings.

\textbf{Kinetics-700.} We also train a model on the Kinetics-700 dataset to serve as a reference model \citep{smaira2020}. Kinetics-700 is a large-scale dataset of short YouTube clips. It contains approximately 536K video clips (training set) representing 700 fine-grained action categories. The clips in Kinetics-700 last less than 10 seconds on average (8.9 seconds per clip) for a total of 1330 hours of video. Thus, compared to the HVM-1 pretraining data, Kinetics-700 is much more diverse in content and has a much shorter time scale.

\textbf{Preprocessing.} We temporally subsample the videos at a rate of 3.75 frames/second and feed the models a stack of 16 consecutive frames at a spatial resolution of either 224$\times$224 or 448$\times$448 pixels. The models thus ``see'' roughly \colorbox{lime}{4.3 second long} clips during both training and evaluation, therefore temporal regularities at longer time scales are beyond the capacity of the models to learn.  

\subsection{Self-supervised learning algorithm}
\textbf{Algorithm.} We use the \colorbox{lime}{spatiotemporal extension of masked autoencoders} \citep{feichtenhofer2022} as our self-supervised video representation learning algorithm. In spatiotemporal MAEs, an input clip is divided into three-dimensional spatiotemporal ``patches'', or boxes. During self-supervised training, a large fraction of these boxes are randomly masked out and the objective of the algorithm is to predict these masked patches at the pixel level from high-level representations of the visible patches passed through an encoder. We use a masking ratio of 90\% for models trained at the 224$\times$224 spatial resolution and a masking ratio of 95\% for models trained at the larger 448$\times$448 resolution.

\textbf{Pretraining details.} Using a highly \href{https://github.com/eminorhan/optimized-stmae}{optimized implementation} incorporating a combination of FlashAttention-2 \citep{dao2023}, JIT-compilation through \texttt{torch.compile}, fused \texttt{AdamW} optimizer, mixed precision training, distributed data parallelism, and selective decoding of videos, we were able to achieve a high training throughput, \textit{e.g.}~completing over 160 epochs of training on Kinetics-700 (with an effective batch size of 256) on four H100 GPUs in about a week.

We pretrain three models in total: two models on the HVM-1 pretraining data (one at 224$\times$224 resolution and one at 448$\times$448 resolution) and one model on the Kinetics-700 dataset (at 224$\times$224 resolution). We train all three models until convergence (Figure~\ref{pretraining_fig}b), using a stepwise learning rate schedule. We train the Kinetics-700 model for 151 epochs with a learning rate of 0.0001 and then for another 11 epochs with a learning rate of 0.00001. We train the HVM-1 model at the 224$\times$224 resolution for 41 epochs with a learning rate of 0.0001 and then for another 14 epochs with a learning rate of 0.00001. Finally, we train the HVM-1 model at the 448$\times$448 resolution for 85 epochs with a learning rate of 0.0001 and then for another 11 epochs with a learning rate of 0.00001. Due to the longer average duration of the video files in the HVM-1 pretraining data, we utilized ``repeat augmentation'' \citep{feichtenhofer2022} while pretraining the HVM-1 models: each video file within a mini-batch was sampled 16 times for the HVM-1 model trained at the 224$\times$224 resolution and 4 times for the HVM-1 model trained at the 448$\times$448 resolution. Further training details can be found at the accompanying GitHub repository.

\subsection{Model architecture}
In all our experiments, we use a standard 633M-parameter spatiotemporal vision transformer encoder \citep{dosovitskiy2020}, commonly referred to as ViT-H, with a spatiotemporal patch size of 2$\times$14$\times$14, \textit{i.e.}~2 frames in the temporal dimension and 14$\times$14 pixels in the spatial dimensions. For an input clip of size 16$\times$224$\times$224 (that is, 16 frames at a spatial resolution of 224$\times$224 pixels), this yields a total of 8$\times$16$\times$16=2048 spatiotemporal patches (``tokens''), and for an input clip of size 16$\times$448$\times$448, it yields a total of 8$\times$32$\times$32=8192 patches.

\subsection{Evaluation}
Since our main goal is to evaluate the strength of the visual representations learned from human-like video data with a modern self-supervised video representation learning algorithm, we primarily use \colorbox{lime}{\textit{few-shot} supervised finetuning} tasks for quantitative evaluation. More specifically, we use the same downstream evaluation tasks as in \cite{orhan2024b}: 
\begin{itemize}
    \item \textbf{Action recognition:}~10-shot and 50-shot action recognition on the Kinetics-700 and Something-Something-V2 (SSV2) \citep{goyal2017} benchmarks. In the Kinetics-700 benchmark, the 10-shot and 50-shot conditions respectively use 17 and 87 hours of video from the training set to finetune the models. In the SSV2 benchmark, on the other hand, the 10-shot and 50-shot conditions use approximately 2 hours and 9 hours of video, respectively, to finetune the models.
    \item \textbf{Object recognition:}~$\sim$25-shot object recognition on the ImageNet \citep{russakovsky2015} and OOD ImageNet \citep{geirhos2021} benchmarks. For these benchmarks, we use a randomly sampled 2\% subset of the ImageNet training set to finetune the models. This corresponds to 25-26 labeled examples per class on average. As in \cite{orhan2024b}, for these image-based tasks, we repeat each input image 16 times, thus effectively creating a 16-frame static video clip for each image. Although somewhat inefficient, this scheme allows us to use our pretrained video models without any changes in their architecture.
\end{itemize}
Further details about the evaluation tasks and supervised finetuning can be found in \cite{orhan2024b}. By using few-shot supervised evaluation tasks (with only tens of labeled examples per class provided in each task), we aim to make sure that the models' visual representations are learned primarily through self-supervised learning over the pretraining datasets and not through supervised data from the downstream tasks. This also makes our evaluation tasks psychologically more realistic, since children are generally thought not to receive large amounts of semantically labeled data during their early development. For similar considerations of psychological plausibility, we also do not use heavy data augmentation or regularization methods commonly employed in finetuning vision transformer models to prevent overfitting \citep{he2022,feichtenhofer2022}. Such heavy data augmentation or regularization methods make the data less human-like. We instead use only random resized crops and horizontal flips in the spatial domain as our data augmentation methods during finetuning.

\section{Results}
\subsection{Few-shot action recognition}
\begin{figure}
  \centering
	\includegraphics[width=0.95\textwidth, trim=2mm 2mm 2mm 2mm, clip]{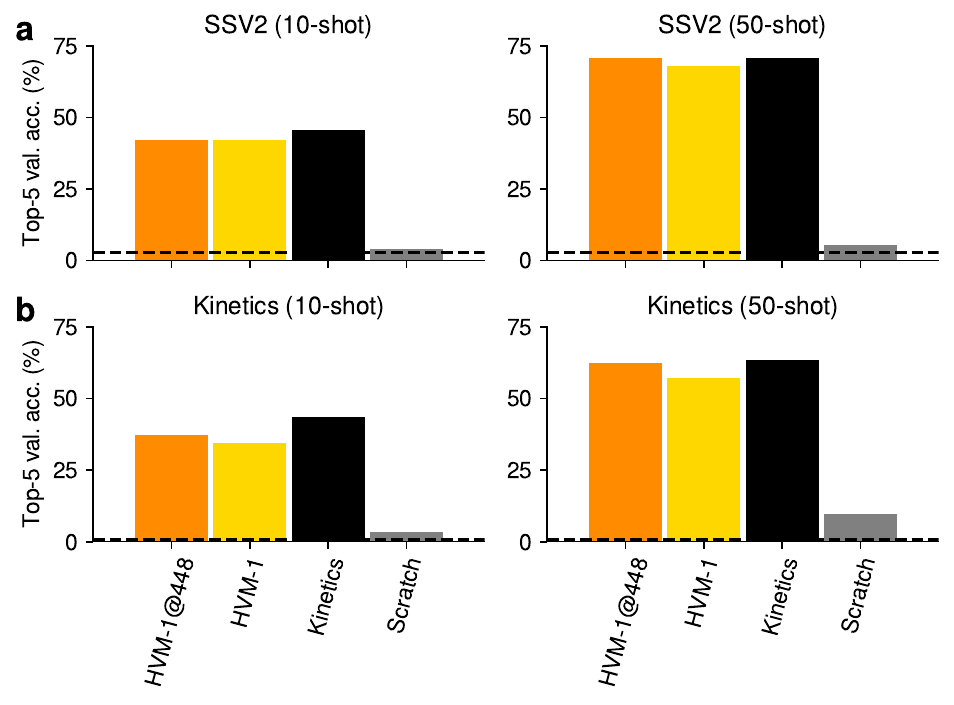} 
	\caption{Top-5 validation accuracy on the few-shot SSV2 \textbf{(a)} and the few-shot Kinetics \textbf{(b)} benchmarks. Model legend: `HVM-1@448' and `HVM-1' are the models pretrained with human-like video data at the 448$\times$448 and 224$\times$224 pixel resolution, respectively; `Kinetics' denotes the model pretrained on the Kinetics-700 training data; `Scratch' refers to a model trained on the downstream task only (with no pretraining). Dashed horizontal lines indicate the chance level.}
  \label{action_fig}
\end{figure}

Figure~\ref{action_fig} shows the top-5 validation accuracy of the models on the few-shot SSV2 (Figure~\ref{action_fig}a) and the few-shot Kinetics (Figure~\ref{action_fig}b) benchmarks (numerical values are documented in Supplementary Table~\ref{numvalues_tab} at the end of the paper). All pretrained models perform substantially better than a model trained from scratch on the downstream task only without any pretraining (denoted as `Scratch' in Figure~\ref{action_fig}). On the SSV2 benchmark, the model pretrained on Kinetics-700 performs slightly better than the HVM-1 model trained at the same spatial resolution (224$\times$224 pixels). The difference between these two models is somewhat greater on the Kinetics benchmark, which is expected, since this benchmark is ``within-distribution'' for the Kinetics pretrained model. The HVM model pretrained at the larger 448$\times$448 resolution (HVM-1@448) reduces this gap and performs competitively with the Kinetics-pretrained model in most cases. We thus conclude that HVM-1 models perform quite well in novel downstream few-shot action recognition tasks and they are competitive against a model pretrained with a large and diverse set of action videos in many cases, despite the fundamentally different nature of the videos these models were pretrained with (temporally extended, continuous, human-like videos \textit{vs.}~short, diverse YouTube clips).

\subsection{Few-shot recognition of a developmentally realistic set of action categories}
\begin{figure}
  \centering
	\includegraphics[width=1\textwidth, trim=0mm 0mm 0mm 0mm, clip]{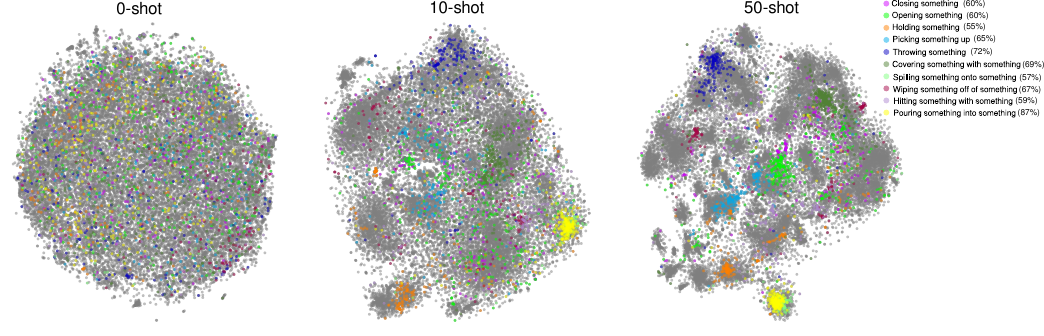} 
	\caption{t-SNE embeddings of the videos from the SSV2 validation set. From left to right, the embeddings were obtained from the pretrained HVM-1@448 model without any finetuning (0-shot), from the pretrained HVM-1@448 model finetuned on the 10-shot SSV2 task (10-shot), and from the pretrained HVM-1@448 model finetuned on the 50-shot SSV2 task (50-shot). Videos belonging to 10 developmentally realistic action categories (listed in the legend) are highlighted with different colors. Gray dots represent the videos belonging to other categories. Numbers in parentheses in the legend represent the top-5 accuracy for the corresponding categories in the 50-shot condition.}
  \label{embeddings_fig}
\end{figure}

We next zoom in on the behavior of the HVM-1@448 model on 10 developmentally realistic action categories from SSV2 (also listed in the legend of Figure~\ref{embeddings_fig}): \textit{closing something}, \textit{opening something}, \textit{holding something}, \textit{picking something up}, \textit{throwing something}, \textit{covering something with something}, \textit{spilling something onto something}, \textit{wiping something off of something}, \textit{hitting something with something}, \textit{pouring something into something}. These are simple, basic action categories that a 2.5-year old child would be expected to understand (see \cite{orhan2024b} for a more detailed description of how exactly these categories were selected). The model performs slightly worse on this developmentally realistic subset compared to the remaining categories: \textit{e.g.}~in the 50-shot condition, 65.1\% top-5 accuracy on the developmentally realistic subset \textit{vs.}~70.8\% on the rest. Similar to \cite{orhan2024b}, we hypothesize that this is likely due to the fact that the developmentally realistic subset contains more generic, more ``abstract'', broader action categories, whereas the remaining categories are, on average, more specific and detailed (for example, specifying a mode of action), which could make them easier to learn. Figure~\ref{embeddings_fig} highlights the t-SNE embeddings of the developmentally realistic action categories for three different versions of the HVM-1@448 model. Embeddings from the 0-shot model that was not finetuned on SSV2 do not display a strong semantic category structure, but the semantic structure readily emerges when the model is finetuned in the 10-shot SSV2 task and is further strengthened in the 50-shot condition.

\subsection{Scaling of action recognition performance with pretraining data size}
\begin{figure}
  \centering
	\includegraphics[width=1\textwidth, trim=2mm 2mm 2mm 2mm, clip]{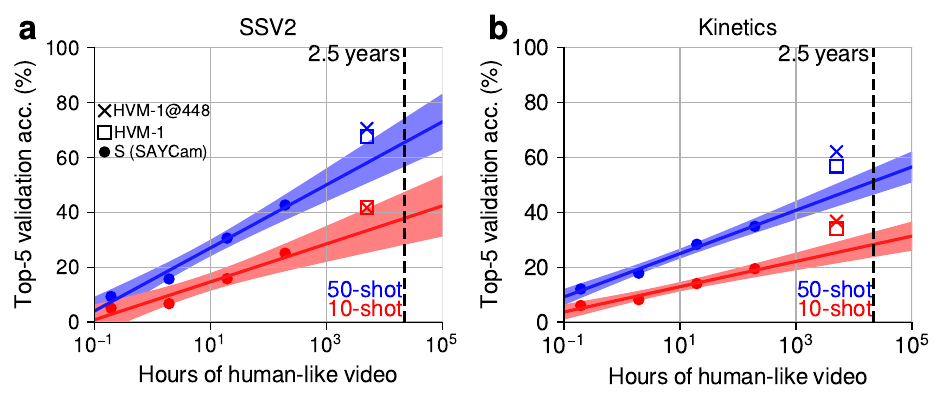} 
	\caption{HVM-1 models ($\square$ and $\times$) outperform the scaling trends estimated from a subset of SAYCam in \cite{orhan2024b}, represented by the solid dots ($\bullet$) and the corresponding log-linear fits. The shaded regions indicate the 95\% confidence intervals around the log-linear fits. Results are shown for the SSV2 \textbf{(a)} and Kinetics-700 \textbf{(b)} benchmarks and for both 10-shot (red) and 50-shot (blue) conditions.}
  \label{scaling_fig}
\end{figure}

\cite{orhan2024b} trained a self-supervised video model on approximately 200 hours of headcam data from a single participant in SAYCam (child S). Since this amount of visual experience is roughly two orders of magnitude smaller than the typical amount of visual experience a developing child would be expected to receive over a few years of early development, \cite{orhan2024b} performed a scaling experiment to extrapolate the action recognition performance of video models beyond this limited amount of available pretraining data. Since the HVM-1 models in this paper are trained with a much larger set of human-like video data, we wanted to revisit the scaling trends estimated in \cite{orhan2024b} in the context of the performance of HVM-1 models. In Figure~\ref{scaling_fig}, the solid dots ($\bullet$) and the log-linear fits show the scaling trends estimated in \cite{orhan2024b} using subsets of data from child S in SAYCam. In both SSV2 (Figure~\ref{scaling_fig}a) and Kinetics-700 (Figure~\ref{scaling_fig}b), the HVM-1 models ($\square$ and $\times$) significantly outperform these previously estimated scaling trends. This could be because of the greater diversity of the HVM-1 dataset (since the dataset contains data from multiple sources and from multiple individuals, whereas the scaling trends in \cite{orhan2024b} were estimated with data from a single participant in SAYCam), or it could be due to other, more subtle differences between the pretraining configurations of \cite{orhan2024b} and the current study. Since estimating new scaling trends from the HVM-1 dataset itself is computationally expensive, we leave a more complete discussion of this topic to future work.

\subsection{Few-shot object recognition}
\begin{figure}
  \centering
	\includegraphics[width=0.95\textwidth, trim=2mm 2mm 2mm 2mm, clip]{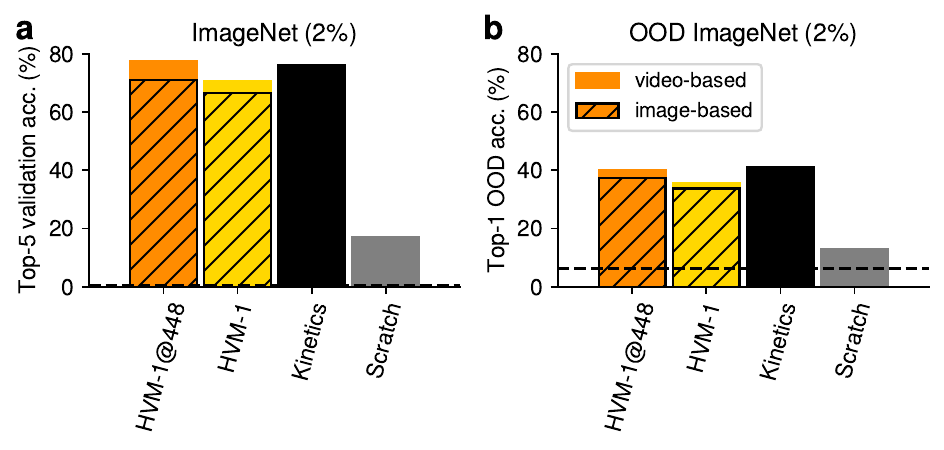} 
	\caption{\textbf{(a)} Top-5 validation accuracy on ImageNet. \textbf{(b)} OOD accuracy on the OOD ImageNet benchmark. Model legend: `HVM-1@448' and `HVM-1' are the models pretrained with human-like video data at the 448$\times$448 and 224$\times$224 pixel resolution, respectively; `Kinetics' denotes the model pretrained on the Kinetics-700 training data; `Scratch' refers to a model trained on the downstream task only (with no pretraining). Dashed horizontal lines indicate the chance performance. Hatched bars represent the performance of image-based models trained with the image-based MAE algorithm on the same data, at the same spatial resolution, and with the same encoder architecture as the corresponding HVM-1 models. These models thus help us isolate the effect of image-based \textit{vs.}~video-based pretraining on object recognition accuracy by controlling for the other main factors.}
  \label{object_fig}
\end{figure}

Finally, Figure~\ref{object_fig} shows the performance of the models on the ImageNet (Figure~\ref{object_fig}a) and OOD ImageNet (Figure~\ref{object_fig}b) object recognition benchmarks (numerical values are documented in Supplementary Table~\ref{numvalues_tab} at the end of the paper). As in the action recognition benchmarks, the model pretrained with the Kinetics-700 dataset slightly outperforms the HVM-1 model trained at the same spatial resolution (224$\times$ 224 pixels), but the HVM-1 model trained at the higher resolution (HVM-1@448) closes this gap and performs comparably to the Kinetics pretrained model.

On both benchmarks, HVM-1 models perform better than image-based models trained with the image-based MAE algorithm \citep{he2022} on the same data, at the same spatial resolution, and with the same architecture as the corresponding HVM-1 models (compare the overlaid solid \textit{vs.}~hatched bars in Figure~\ref{object_fig}). These image-based models were previously introduced in \cite{orhan2023}. The superior performance of video-based models provides suggestive evidence that by learning to predict temporal regularities over short video clips, video-based pretraining may lead to better (\textit{i.e.}~more accurate and more robust) object representations than purely image-based pretraining, which ignores all temporal information. Although this comparison controls for the effects of dataset, spatial resolution, and model architecture, there were other (more minor) differences between the pretraining configurations of the video-based HVM-1 models introduced in this paper and the image-based MAE models introduced in \cite{orhan2023}\footnote{\textit{e.g.}~learning rate schedules and batch sizes were slightly different; masking ratios were chosen based on previous work and were not optimized in either case.}, so we cannot attribute the effect to the video-based \textit{vs.}~image-based distinction with high confidence and we once again leave a more rigorous analysis of this intriguing result to future work. This result also suggests that, using video-based pretraining instead of image-based pretraining, it may be even easier than previously estimated by \cite{orhan2023} to reach human-level object recognition capabilities from human-like video data by simultaneously scaling up the data size, model size, and image resolution.

\section{Discussion}
In this paper, we introduced HVM-1, large scale self-supervised video models trained with nearly 5000 hours of human-like video data. Our results demonstrate the potential of large-scale human-like video data for learning high-quality visual representations despite the fundamentally different spatiotemporal characteristics of such data from the type of visual data modern machine learning models are usually trained with. 

We hope that these models will be a useful tool especially for people interested in understanding the similarities and differences between humans and machine learning models in the visual domain. However, a few important caveats are in order before making such comparisons:
\begin{itemize}
    \item Though human-like in the ways discussed earlier in the Methods section, the HVM-1 dataset does not represent the visual experiences of a single individual (since it contains data from multiple sources and from multiple individuals), so there still remains a gap between the HVM-1 dataset and truly representative human-like visual experience.\footnote{For obvious practical reasons, this gap is very challenging to close entirely and will likely remain so for the foreseeable future.} It is unclear how consequential this gap is. Relatedly, even though the HVM-1 dataset constitutes the largest human-like video dataset to date that we know of, it still contains only about half a year of visual experience (or a little over a year of visual experience if we factor in 12 hours/day of sleep). This is still about an order of magnitude less visual experience than the amount a typical mature human being experiences over 10-20 years in their early life.
    \item The HVM-1 models do not aim to be biologically (or psychologically) plausible with respect to their architecture or learning algorithm. The human brain is subject to idiosyncratic constraints the HVM-1 models are not (and \textit{vice versa}). The HVM-1 models rather aim to be similar to humans with respect to the data distribution they learn from.
    \item The HVM-1 models learn from \textit{visual experience only}, whereas humans learn from an inherently embodied and multimodal sensory stream, which probably profoundly affects the character of visual experiences in humans.
\end{itemize}

\section*{Acknowledgements}
I am grateful to the creators and maintainers of the datasets used in this study. I would also like to thank NYU HPC for providing the compute resources.

\bibliography{scaling}
\bibliographystyle{apalike}

\clearpage
\renewcommand{\tablename}{Supplementary Table}
\begin{table}
\centering
\begin{tabular}{cccccc}
 Task & HVM-1@448 & HVM-1 & Kinetics & Scratch & Figure reference\\[1mm]
\hline
 SSV2 (10-shot) & 41.7 & 41.8 & 45.1 & 3.6 & Figure~\ref{action_fig}a \\
 SSV2 (50-shot) & 70.5 & 67.7 & 70.4 & 5.1 & Figure~\ref{action_fig}a \\
 Kinetics (10-shot) & 36.8 & 34.1 & 43.3 & 3.1 & Figure~\ref{action_fig}b \\
 Kinetics (50-shot) & 62.1 & 56.8 & 62.9 & 9.2 & Figure~\ref{action_fig}b \\
 ImageNet (2\%) & 77.3 (V) / 71.0 (I) & 70.7 (V) / 66.5 (I) & 76.1 & 17.0 & Figure~\ref{object_fig}a \\
 OOD ImageNet (2\%) & 40.2 (V) / 37.4 (I) & 35.6 (V) / 33.8 (I) & 41.0 & 12.8 & Figure~\ref{object_fig}b \\
\end{tabular}
\caption{Numerical values for the results reported in Figures~\ref{action_fig} and \ref{object_fig}. For the ImageNet and OOD ImageNet benchmarks in Figure~\ref{object_fig}, we report the performance of both the video-based (V) and the image-based models (I).}
\label{numvalues_tab}
\vspace{0em}
\end{table}

\end{document}